\documentclass[10pt,twocolumn,letterpaper]{article}

\usepackage{cvpr}
\usepackage{times}
\usepackage{epsfig}
\usepackage{graphicx}
\usepackage{amsmath}
\usepackage{amssymb}
\usepackage{cvpr}
\usepackage{times}
\usepackage{epsfig}
\usepackage{graphicx}
\graphicspath{ {../images/} }
\usepackage{amsmath, amsfonts}
\usepackage{amssymb}
\usepackage{caption}
\usepackage{svg}
\usepackage{tabularx,threeparttable, booktabs}
\usepackage{mathtools}

\DeclarePairedDelimiter\floor{\lfloor}{\rfloor}

\newcommand*\samethanks[1][\value{footnote}]{\footnotemark[#1]}

\usepackage{hyperref}
\hypersetup{breaklinks=true,bookmarks=false}

\cvprfinalcopy

\def\cvprPaperID{****} 

\begin{document}

\title{BEV-Seg: Bird's Eye View Semantic Segmentation Using Geometry and Semantic Point Cloud}

\author{Mong H. Ng\thanks{indicates equal contribution},\,
Kaahan Radia\samethanks,\,
Jianfei Chen,\, Dequan Wang,\, Ionel Gog,\, Joseph E. Gonzalez \\
University of California, Berkeley \\
{\tt\small \{monghim.ng, kaahan2021, jianfeic, dequanwang, ionel, jegonzal\}@berkeley.edu}
}
\maketitle

\begin{abstract}

Bird's-eye-view (BEV) is a powerful and widely adopted representation for road scenes that captures surrounding objects and their spatial locations, along with overall context in the scene. In this work, we focus on bird's eye semantic segmentation, a task that predicts pixel-wise semantic segmentation in BEV from side RGB images. This task is made possible by simulators such as Carla, which allow for cheap data collection, arbitrary camera placements, and supervision in ways otherwise not possible in the real world. There are two main challenges to this task: the view transformation from side view to bird's eye view, as well as transfer learning to unseen domains. Existing work transforms between views through fully connected layers and transfer learns via GANs. This suffers from a lack of depth reasoning and performance degradation across domains. Our novel 2-staged perception pipeline explicitly predicts pixel depths and combines them with pixel semantics in an efficient manner, allowing the model to leverage depth information to infer objects' spatial locations in the BEV. In addition, we transfer learning by abstracting high-level geometric features and predicting an intermediate representation that is common across different domains. We publish a new dataset called BEVSEG-Carla and show that our approach improves state-of-the-art by 24\% mIoU and performs well when transferred to a new domain.

\end{abstract}

\section{Introduction}

A perception system that reliably recognizes the locations and types of surrounding objects is a key component of autonomous driving systems. Perception is typically achieved with various sensors, including RGB cameras, thermal cameras, lidar, and radar. While lidar provides the most accurate depth estimation, the expensive cost hinders its mass deployment. Moreover, it is crucial to have a backup system in case of any sensor failures for applications that strongly emphasizes safety. It is therefore desirable to have a perception system that builds purely on camera inputs. In addition, such a perception system will need to output predictions that are convenient to reason with for subsequent autonomous driving subsystems, such as planning and prediction modules. Recently, rasterized bird's-eye-view (BEV) representations of the world have gained popularity as a representation that captures the environment in a suitable manner for planning and prediction \cite{bansal2018chauffeurnet,wang2019monocular,chen2019learning,intentnet}. A BEV representation's grid-structured representation readily allows for the direct application of convolutional layers and efficient inference of neural networks.

With these two requirements in mind, the task of image-based bird's-eye semantic segmentation is as followed: given $N$ images capturing the road scene from an AV at different angle, predict a pixel-wise semantic segmentation $B$ in the bird's-eye view. Data collection for this task is made possible by simulators such as GTA-V \cite{gtav} and CARLA \cite{carla}, which allow for inexpensive data collection and ground-truth labels otherwise unobtainable in the real world. A recent work VPN tackles this task by using convolutional neural network (CNN) on side view and later again on bird's-eye view,  transforming features from side views to bird's-eye view by fully connected layers \cite{mitpaper}. This approach, while simple, suffers from lack of depth reasoning and results in coarse bird's-eye segmentation. Furthermore, because ground truths can only be collected in the simulator domain, it is imperative to transfer the model to achieve good performance in other target domains. In VPN, this problem is exacerbated by the its end-to-end training. VPN adapts the model by using a GAN approach but achieves much worse performance compared to that of the source domain.


In this work, we present a novel 2-staged perception pipeline for autonomous driving that leverages depth and geometric information to improve BEV segmentation. In stage 1, our pipeline consumes monocular images from various angles, then use a depth estimation module and a semantic segmentation module to predict a depth and segmentation map for each view. We then combine the two into a semantic point cloud. In stage 2, a parser network constructs the bird's-eye-view segmentation by operating over the projected semantic point cloud. We address the lack of depth reasoning by explicitly training a monocular depth estimator as part of our pipeline and transforming pixels into bird's-eye view by geometry and pin-hole camera model. We address the problem of transfer learning by using an intermediate representation between stages and abstracting away features that are common across domains. We train our model with RGB images, segmentation and depth ground truths in the side view, and BEV ground truth semantic segmentation collected in the CARLA simulator \cite{carla}. Compared with previous pipeline \cite{mitpaper}, our pipeline raises segmentation mIoU from 36.4\% to 60.4\%. Thanks to the extra level of intermediate representation, our approach also achieves better transfer performance between different CARLA environments. 

In summary, our contributions include:
\begin{enumerate}
\itemsep0em
    \item We propose a perception pipeline that incorporates depth information, geometry, and semantics in a novel way to predict bird's-eye-view segmentation;
    \item We obtain the new state-of-the-art results in bird's-eye-view segmentation, outperforming the previous state-of-the-art by $20-30$\% mIoU;
    \item We propose an alternate intermediate representation that is more robust and agnostic to sensor type and environment, and show that this representation allows for better transfer learning between different domains; and
    \item We release a new CARLA dataset with rich variation in the weather, landscape, and surrounding vehicle domains for use in studying domain adaptation in autonomous driving settings.
\end{enumerate}

\begin{figure}
    \centering
    \includegraphics[width=\linewidth]{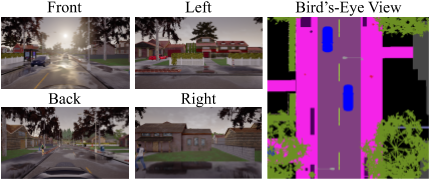}
    \caption{Example of RGB, BEV dataset pair.}
    \label{fig:datasetexample}
\end{figure}

\begin{figure*}[t]
    \includegraphics[width=\textwidth]{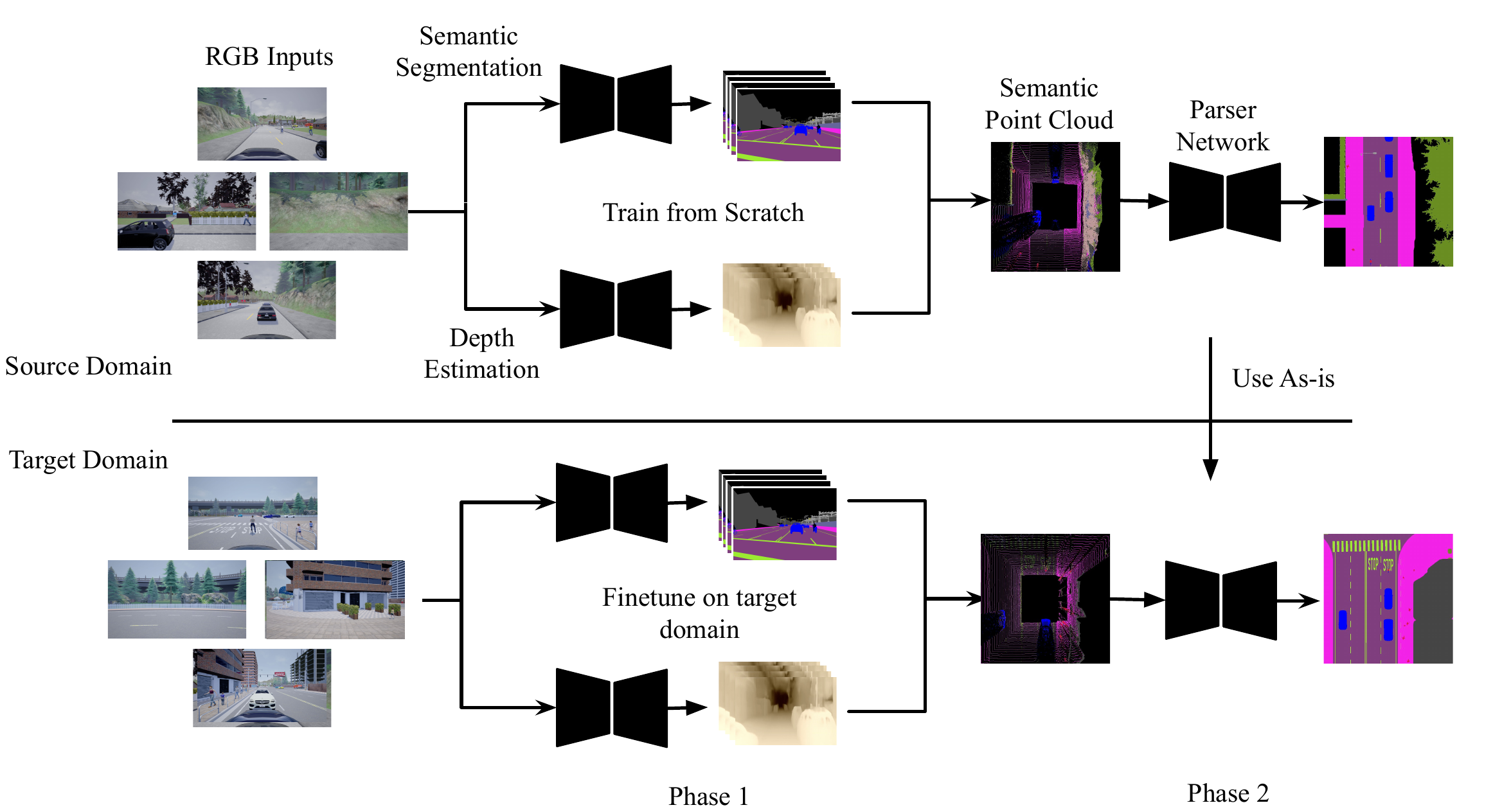}
\caption{The proposed BEV-Seg pipeline.}
\label{fig:pipeline}
\end{figure*}
\section{Related Works}

\subsection{Bird's-Eye View Representations}

Predicting the semantic layout of a scene is a well-studied task in computer vision \cite{clutteredroom, im2cad, pano2cad, layoutnet, roomnet}, usually trained using geometric constraints or full annotations of the layout (e.g., dense ground truths). While our method, like Layoutnet \cite{layoutnet}, constructs a 3D layout of the scene, the layout (a semantic point cloud) is purely an intermediate form to reduce computational complexity for the scene semantics. Several works \cite{ganbev, geometricbev, Lin_2012} have pushed the boundary for view synthesis, which generates realistic views of a scene from different view points. Our work is concerned with generating a semantic representation of the bird's eye view. MVP \cite{wang2019monocular} predicts a bird's eye view by detecting 3D objects, but it disregard other features such as roads, road lanes, and buildings. Recently, VPN \cite{mitpaper} proposes the task of bird's-eye view semantic segmentation, using inputs and ground truth collected from the CARLA simulator. It generates the BEV segmentation map in an end-to-end fashion using a fully connected layer to transform side-view images to bird's eye view. This approach, however, fails to leverage geometry and depth information. Its end-to-end training also makes the model less transferable to new domains.

\subsection{Image-Derived Depth}

Depth information is important in autonomous vehicles as objects need to be reasoned in the correct locations for obstacle avoidance. Lidar sensors provide the most accurate measurements in depth, but are expensive and its sparsity provides incomplete depth information. 3D object detection from images \cite{chen2017multi,yoo20203d} predict the location and size of important object such as vehicles, pedestrians, and cyclists. MVP \cite{wang2019monocular} extends these 3D object detection models by mapping these objects in BEV, generate a rasterized BEV image, and construct a driving policy. While these object-based depth predictions are light-weight and accurate in predicting objects, they ignore other features such as road lanes and signs that also important in scene understanding. In addition, depth estimation predicts the depth of every pixel using monocular images, while stereo matching find correspondence between pixels to predict disparity, which can then be converted into pixel-wise depths. Many works combine pixel-wise depth maps with semantic features by channel-wise concatenation \cite{mitpaper,stereoaccurateobjectdetection,robustobject}. In contrast, pseudolidar-based approaches \cite{pseudolidar,you2019pseudolidarpp,qian2020endtoend,vianney2019refinedmpl} convert depth maps into the format of point cloud and show that this is a easier representation for neural network to process. In this work, not only do we predict a pseudolidar using monocular depth estimation, we also concatenate each point with the predicted classes from semantic segmentation to form a semantic point cloud.

\subsection{Transfer Learning}

Given the large domain gap between different driving environments, many different methods were developed to transfer learning between them. \cite{transferablefeatures, coupledgans, crossdomainimagegen, pixeldomain, dcan} implement domain adaption techniques to mitigate the large domain gap between synthetic images and real world images. \cite{geodesic1, geodesic2} both augment this using geodesic distances to match feature distributions. VPN \cite{mitpaper} builds on the aforementioned works and bridges the domain gap in the context of BEV semantic segmentation, using adversarial networks to match distributions of the source domain and target domain. While transferred model achieve qualitatively meaningful results, the outputs are coarser and less accurate.

Instead of using an end-to-end network and then applying a transfer learning component, some works use a modular pipeline and train different modules on different domains. \cite{artags} trains two neural networks: one that identifies an object's position, and another that takes as input the object pose and outputs actions. To transfer from the simulator to the real world they train a new object pose network and use the action network as-is. \cite{modnueral} trains multiple networks of the same architecture for robots doing different tasks, and transfer learning across task by swapping those trained networks. \cite{mullerpaper} is closely related to our conceptualization of a modular AV pipeline. It uses a binary semantic segmentation of their frontal view as an intermediate representation; though this is useful for transfer learning, it falls short in utility for downstream tasks due to its lack of detailed semantic and depth information. Similarly, our pipeline uses semantics in a semantic point cloud, which incorporates both depth and semantic information. However, our immediate representation is the projection of semantic point cloud onto a rasterized image, thus not only abstracting the semantics but also the depth information by putting continuously valued depths into rasterized bins.

\begin{figure*}
    \centering
    \includegraphics[width=\linewidth]{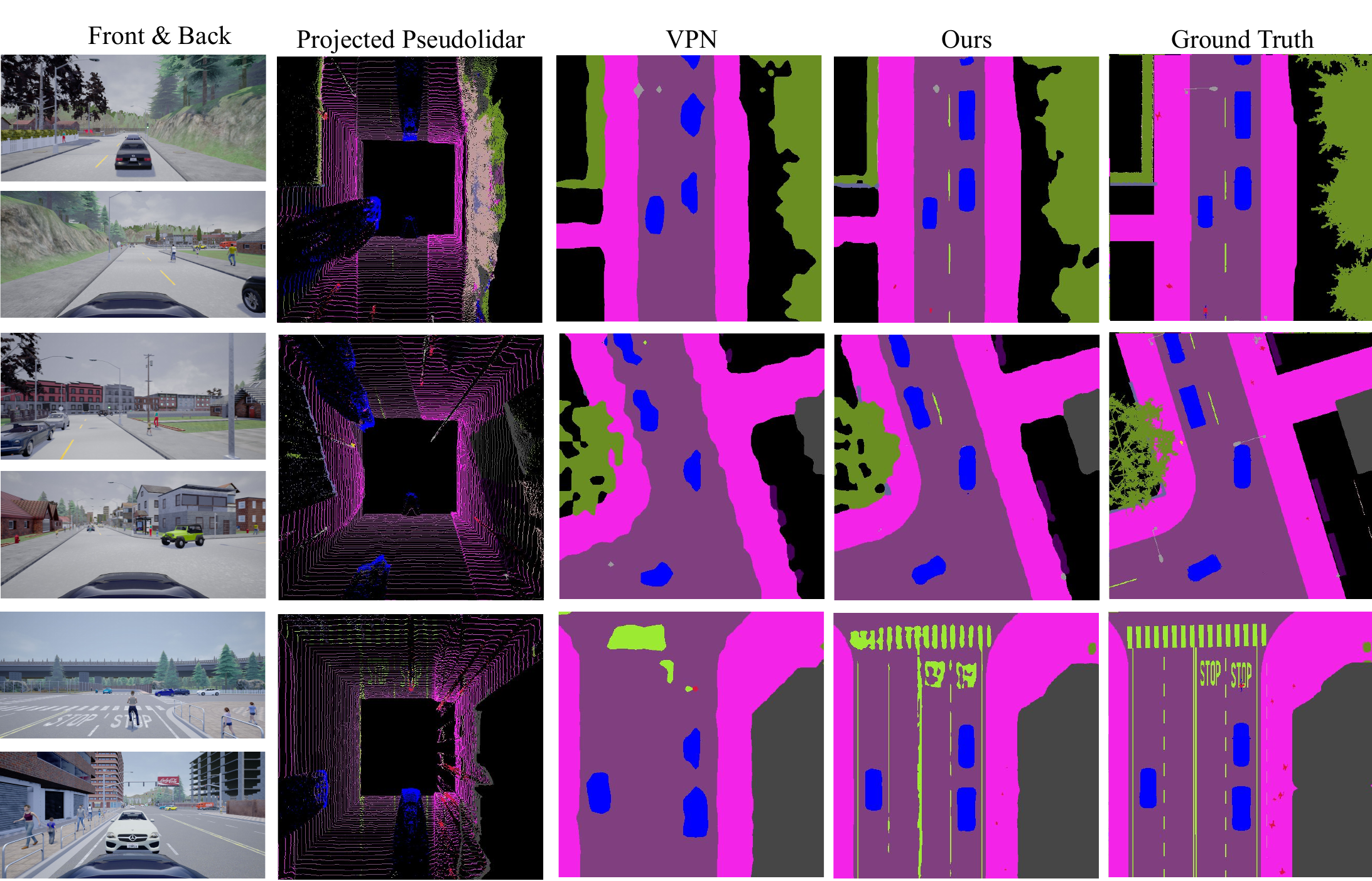}
    \caption{Qualitative Results on the BEVSEG-Carla dataset.}
    \label{fig:qualitative_result}
\end{figure*}

\begin{table*}
    \begin{center}
        \begin{tabular}{|c | c|}
            \hline
            Model & mIoU \\
            \hline\hline
            VPN RGB & 36.4\% \\ 
            VPN RGB-D & 37.3\%  \\ 
            \hline
            ours & 60.4\% \\
            ours - Segmentation Oracle & 60.8\% \\ 
            ours - Depth Oracle & 66.5\% \\ 
            ours - Depth \& Segmentation Oracle & 67.3\% \\ 
            \hline
        \end{tabular}
    \end{center}
\caption{Segmentation Result on  BEVSEG-Carla. Oracle models have ground truth given for specified inputs.}

\label{table:abalation}
\end{table*}

\section{Methodology}

\subsection{A Two-Stage Perception Pipeline}

Our pipeline consists of two stages, as depicted in Figure \ref{fig:pipeline}. In the first stage, $N$ RGB road scene images are captured by cameras at different angles and individually pass through semantic segmentation network $S$ and depth estimation network $D$. The resulting side semantic segmentations and depth maps are combined and projected into a semantic point cloud. This point cloud is then projected downward into an incomplete bird's-eye view, which is fed into a parser network \cite{hrnet} to predict the final bird's-eye segmentation. The rest of this section provides details on the various components of the pipeline.

\paragraph{Stage 1: RGB to Intermediate Representation}
Given $N$ RGB images captured at different angles, we pass each into a semantic segmentation network and a monocular depth estimation network. Our semantic segmentation network $S$ takes in side RGB images $I_l$, $l \in \{1, \cdots, N\}$ and produces a segmentation map $S_l$. Given $C$ output classes (pedestrian, car, etc.), our segmentation map is represented as $S_{l}(u, v) = c_{l, u, v}$ where $u, v$  are the pixel location and $c_{l, u, v}$ is the predicted class for that image and location. Our monocular depth estimation network $D$ also takes in side RGB images $I_l$ and produces depth maps $D_l(u, v) = d_{l, u, v}$ where each pixel location $u, v$ contains a predicted continuous depth value $d_{l, u, v}$. The semantic segmentation network is supervised by ground truth segmentation labels, which could be human-annotated or simulator generated. The depth estimation network is supervised by ground truth lidar projected onto the image plane.

For each view $l$, we then create a \textit{semantic} point-cloud $\mathcal{P}_l$ from its corresponding segmentation map $S_l$ and depth map $D_{l}$. To do this, we first combine $S_l$ and $D_l$ to obtain a semantic perspective point cloud $\{(u^{(i)}, v^{(i)}, d^{(i)}, c^{(i)})\}_{i = 1}^{K_l}$, where $K_l$ is the number of pixels from view $l$. Using the pinhole camera model, we project each perspective point into $3D$ camera coordinates by the following equations:

\begin{align*}
z^{(i)} &= d^{(i)},\\
x^{(i)} &= \frac{(u^{(i)} - c_U) \times z}{f_U},\\
y^{(i)} &= \frac{(v^{(i)} - c_V) \times z}{f_V},
\end{align*}

where $f_U, f_V$ are the horizontal and vertical focal lengths, and $(c_U, c_V)$ is the pixel center of the image. This results in a view specific point cloud $P_l = \{(x^{(i)}, y^{(i)}, z^{(i)}, c^{(i)})\}_{i = 1}^{K_l}$. We then rotate each point cloud into its respective view in the 3D vehicle coordinates using rotation matrices. Concatenating all point clouds, we obtain a final semantic point cloud $P = \{(x^{(i)}, y^{(i)}, z^{(i)}, c^{(i)})\}_{i = 1}^{K}$, where $K$ is the total number of pixels from all views. Finally, we remove the height dimension $y$ by orthographically projecting the points downward onto an incomplete bird's-eye segmentation $T$ of size $H_{BEV} \times W_{BEV}$. For each point $(x^{(i)}, y^{(i)}, z^{(i)}, c^{(i)})$ we set the pixel value $T(\floor{x^{(i)}}, \floor{y^{(i)}}) = c^{(i)}$. We resolve projection conflicts by always choosing the point of lower height, since the most important obstacles for driving are often near the ground. We assign pixels where there is no corresponding point the semantic label of \textit{void}. This is the final intermediate representation that bridges stage 1 and stage 2.

\paragraph{Stage 2: Intermediate Representation to Bird's-Eye Semantic Segmentation}
We expand the incomplete bird's-eye segmentation $T$ into one-hot encoding along the class $c$ dimension. This results in a tensor $L$ of size $H \times W \times (C + 1)$, where $H, W$ are the height and width of $T$, and $C$ is the number of classes with the additional class of \textit{void} class. This is passed into a parser network to predict the full bird's-eye semantic segmentation.

Our parser network is another semantic segmentation network $P$. The purpose of the parser network is to take the incomplete bird's-eye view and "fill" in the void pixels as well as locally "smooth" already predicted segmentation locations through convolution operations. Thus we take our incomplete bird's-eye view tensor $L$ and pass it through $P$ to generate our final BEV segmentation map $B$ of size $H \times W$. Void is not one of the predicted class as they are recovered by the parser network.

\paragraph{Advantages of Explicit Depth Reasoning and Using Geometry to Transform Between Views}

In our modular approach, each network can be trained separately and be enforced to reason in specific space. The side view segmentation network is tasked with reasoning with the semantics of each pixels, while the depth network is tasked with reasoning with the locations of each pixel. In the baseline VPN \cite{mitpaper}, depth information is reasoned by fully connected layers and supervised by training signals from the ground truth bird's-eye view. In our approach, however, we have a separate module for depth reasoning and also supervise it with ground truth depth maps, providing stronger depth training signals. Our choice of intermediate representation enables meaningful transformation from side view to BEV. First, pin-hole camera model alleviates the inherent perspective warp in image plane by correcting points to their correct 3D locations. Most importantly, by using this geometry, we bring together objects that are far in the image plane but close in the 3D space. On an image, background buildings are closer to vehicles and thus reasoned closer together, whereas on BEV, vehicles' 3D locations are recovered and are reasoned closer along the road instead.

\subsection{Transfer Learning Via a Common Intermediate Representation}

Transfer learning is important to autonomous vehicles because AV could be deployed in unseen environments, but it is difficult in the task of bird's-eye semantic segmentation. It is much more difficult to collect ground truths for bird's-eye semantic segmentation, without which we cannot fine-tune a model to target domain. Therefore, prior works \cite{mitpaper} employs domain adaptation, but the result is less than ideal. In our pipeline, we propose transfer learning via modularity and abstraction. To transfer from the source domain to a target domain, we

\begin{enumerate}
    \item Fine-tune the stage 1 models on the target domain stage 1 data
    \item Apply the trained stage 2 model as-is to the projected point cloud in the target domain
\end{enumerate}

In stage 1, while a domain gap does exist, the inputs and ground truths (input RGB images, semantic ground truths, and depth ground truths) can be readily collected and annotated in target domain, allowing us to fine-tune. On the other hand, in stage 2, while the ground truth of bird's-eye segmentation is difficult to collect in most domains, the intermediate common representation of projected semantic point cloud reduces the domain gap by abstracting geometric features such as object shapes and road curves, from the more domain-dependent features such as weather condition, time of day, and camera settings. Therefore, we can simply use the same model trained in the source domain. Qualitative and quantitative results in section \ref{experiment_results} demonstrate that this simple scheme achieves good transfer learning performance.

\subsection{Pipeline and Network Configuration}
\label{sec: pipeline config}
For side-semantic segmentations, we use HRNet \cite{hrnet}, a state-of-the-art convolutional network for semantic segmentation. For monocular depth estimation, we implement SORD \cite{sord} using the same HRNet as the backbone. For both tasks, we train the same model on all four views. The resulting semantic point cloud is projected height-wise onto a 512x512 image. We train a separate HRNet model as our parser network for the final bird's-eye segmentation.

\begin{table*}
    \begin{center}
        \begin{tabular}{|c | c c |}
            \hline
            Model & Source Domain & Target Domain after Transfer Learning)\\
            \hline\hline
            VPN~\cite{mitpaper} & 36.4\% & 27.8\% \\ 
            \hline
            ours & 60.4\% & 44.5\% \\
            \hline
        \end{tabular}
    \end{center}
\caption{Results (mIoU) on Transfer Learning from Clear Noon to Wet Sunset}
\label{table:transfer}
\end{table*}

\begin{table*}[h]
    \begin{center}
        \begin{tabular}{|c | c c c c c c c c c|}
            \hline
            Model & Buildings & Fences & Pedestrians & Poles & Road Lines & Roads & Sidewalks & Vehicles & Walls \\
            \hline\hline
            VPN~\cite{mitpaper} & 80.2\% & 11.0\%& 0.7\%& 3.83\%& 5.38\%& 87.8\%& 71.7\%& 53.9\%& 14.6\%\\ 
            \hline
            ours & 89.9\% & 62.3\%& 15.4\%& 33.1\%& 48.2\%& 96.1\%& 91.0\%& 88.3\%& 59.0\%\\
            \hline
        \end{tabular}
    \end{center}
\caption{Class IoU on Source Domain.}
\label{table:ciou}
\end{table*}
\section{Experiment Results} \label{experiment_results}


We collect a dataset of 14k frames in the CARLA simulator by driving a vehicle equipped with various sensors in predefined routes around the designated town. Data collection is as follows: 4 cameras are mounted at the top of the vehicle, each facing in direction perpendicular to each other with a 90 degree field of view, so that together the entire surrounding is captured. At all four angles, we capture RGB road scene images, ground truth semantic segmentations, and ground truth depth maps, all at the resolution of $1024 \times 576$. To capture bird's-eye semantic segmentation, we place an imaginary camera 200m above the ground facing down. The segmentation is captured with the field of view of $8.58$ degree and at the resolution of $256 \times 256$, covering an area of $15m \times 15m$. In 85 episodes, our vehicle drives around 6 towns in 2 weather conditions: clear noon and wet sunset. Samples of the dataset are shown in Figure \ref{fig:datasetexample}. We compare our two-staged pipeline with the baseline VPN \cite{mitpaper}, which trains an end-to-end network for bird's-eye view segmentation.

\paragraph{Benchmarks on Source Domain} In this experiment, the source domain contains frames with the clear noon weather condition. We train two versions of VPN, one with only RGB images as inputs and one with depth maps added. For our approach, in addition to our model that takes as input RGB side images, we also train three other variants of our models with extra inputs: ground truth side segmentations, ground truth depth maps, and both. Table \ref{table:abalation} lists the accuracy of the various models in mIoU on the source domain. We note that for the baseline, after we append a depth channel, accuracy does not improve much. For our approach, note that the last model with both ground truth side segmentation and depth maps, operates exclusively in stage 2, thus removing the error introduced in stage 1 and serving as an oracle to our approach. Replacing ground truth side segmentation with predicted segmentation, we observe a slight decrease in accuracy. This can be thought of as an alternate setting of deploying RGBD sensors. Alternatively, taking as input predicted depth maps but ground truth side segmentation results in a larger decrease in accuracy to 60.8\%, revealing that accurate depth maps are essential for this approach. Lastly, our approach replaces both ground truths with their respective predicted counterparts. We observe another decrease in accuracy to 60.4\%, but our approach still outperforms both the baselines. Qualitatively results are in Figure \ref{fig:qualitative_result}. We note that overall, our approach generates predictions that are sharper and finer in detail.

\paragraph{Benchmarks on Target Domain} 

To test transfer learning, we train on the source domain of clear noon and evaluate on frames in the wet sunset weather condition. Results are listed in Table \ref{table:transfer}. For the baseline, we run its GAN-based domain adaptation and observe a drop in performance. For our pipeline, we retrain stage 1 models and reuse the stage 2 model and perform well on the target domain.

\paragraph{Predicting Important Classes} Table \ref{table:ciou} lists the IoU per class. We note that the baseline is unable to predict many of the important classes, such as pedestrians and cyclists. On the other hand, our approach not only recognizes these important objects, but also recognizes road lanes and stop signs, which are indispensable for later planning stages. In general, for smaller and subtler objects, our approach outperforms the baseline by a large margin.

\section{Conclusion}

We propose a novel solution to the bird's-eye view semantic segmentation problem directly from RGB images. Our two-staged pipeline consists of a side segmentation module and a side depth estimation module to generate semantic pseudolidar, a geometric projection of that point cloud onto a rasterized BEV, and a parser network to recover the full bird's-eye semantic segmentation. We demonstrate that this modular pipeline enables simpler transfer learning with better accuracy in the target domain. We also provide a new Bird's Eye Segmentation Dataset with wider variety and greater difficulty than previously released datasets.

{\small
\bibliographystyle{ieee_fullname}
\bibliography{egpaper_final}

\begin{thebibliography}{10}\itemsep=-1pt

\bibitem{geometricbev}
Ammar Abbas and Andrew Zisserman.
\newblock A geometric approach to obtain a bird's eye view from an image.
\newblock {\em CoRR}, abs/1905.02231, 2019.

\bibitem{bansal2018chauffeurnet}
Mayank Bansal, Alex Krizhevsky, and Abhijit Ogale.
\newblock Chauffeurnet: Learning to drive by imitating the best and
  synthesizing the worst.
\newblock {\em arXiv preprint arXiv:1812.03079}, 2018.

\bibitem{pixeldomain}
Konstantinos Bousmalis, Nathan Silberman, David Dohan, Dumitru Erhan, and Dilip
  Krishnan.
\newblock Unsupervised pixel-level domain adaptation with generative
  adversarial networks.
\newblock {\em CoRR}, abs/1612.05424, 2016.

\bibitem{intentnet}
Sergio Casas, Wenjie Luo, and Raquel Urtasun.
\newblock Intentnet: Learning to predict intention from raw sensor data.
\newblock In Aude Billard, Anca Dragan, Jan Peters, and Jun Morimoto, editors,
  {\em Proceedings of The 2nd Conference on Robot Learning}, volume~87 of {\em
  Proceedings of Machine Learning Research}, pages 947--956. PMLR, 29--31 Oct
  2018.

\bibitem{chen2019learning}
Dian Chen, Brady Zhou, Vladlen Koltun, and Philipp Krähenbühl.
\newblock Learning by cheating, 2019.

\bibitem{stereoaccurateobjectdetection}
Xiaozhi Chen, Kaustav Kundu, Yukun Zhu, Huimin Ma, Sanja Fidler, and Raquel
  Urtasun.
\newblock 3d object proposals using stereo imagery for accurate object class
  detection.
\newblock {\em CoRR}, abs/1608.07711, 2016.

\bibitem{chen2017multi}
Xiaozhi Chen, Huimin Ma, Ji Wan, Bo Li, and Tian Xia.
\newblock Multi-view 3d object detection network for autonomous driving.
\newblock In {\em Proceedings of the IEEE Conference on Computer Vision and
  Pattern Recognition}, pages 1907--1915, 2017.

\bibitem{artags}
Ignasi Clavera, David Held, and Pieter Abbeel.
\newblock Policy transfer via modularity and reward guiding.
\newblock In {\em Proceedings of International Conference on Intelligent Robots
  and Systems (IROS)}, September 2017.

\bibitem{modnueral}
Coline Devin, Abhishek Gupta, Trevor Darrell, Pieter Abbeel, and Sergey Levine.
\newblock Learning modular neural network policies for multi-task and
  multi-robot transfer.
\newblock {\em CoRR}, abs/1609.07088, 2016.

\bibitem{sord}
Raul Diaz and Amit Marathe.
\newblock Soft labels for ordinal regression.
\newblock In {\em Proceedings of the IEEE Conference on Computer Vision and
  Pattern Recognition}, pages 4738--4747, 2019.

\bibitem{carla}
Alexey Dosovitskiy, German Ros, Felipe Codevilla, Antonio Lopez, and Vladlen
  Koltun.
\newblock {CARLA}: {An} open urban driving simulator.
\newblock In {\em Proceedings of the 1st Annual Conference on Robot Learning},
  pages 1--16, 2017.

\bibitem{clutteredroom}
V. {Hedau}, D. {Hoiem}, and D. {Forsyth}.
\newblock Recovering the spatial layout of cluttered rooms.
\newblock In {\em 2009 IEEE 12th International Conference on Computer Vision},
  pages 1849--1856, 2009.

\bibitem{im2cad}
Hamid Izadinia, Qi Shan, and Steven~M. Seitz.
\newblock {IM2CAD}.
\newblock {\em CoRR}, abs/1608.05137, 2016.

\bibitem{roomnet}
Chen{-}Yu Lee, Vijay Badrinarayanan, Tomasz Malisiewicz, and Andrew Rabinovich.
\newblock Roomnet: End-to-end room layout estimation.
\newblock {\em CoRR}, abs/1703.06241, 2017.

\bibitem{Lin_2012}
Chien-Chuan Lin and Ming-Shi Wang.
\newblock A vision based top-view transformation model for a vehicle parking
  assistant.
\newblock {\em Sensors}, 12(4):4431–4446, Mar 2012.

\bibitem{coupledgans}
Ming{-}Yu Liu and Oncel Tuzel.
\newblock Coupled generative adversarial networks.
\newblock {\em CoRR}, abs/1606.07536, 2016.

\bibitem{transferablefeatures}
Mingsheng Long and Jianmin Wang.
\newblock Learning transferable features with deep adaptation networks.
\newblock {\em CoRR}, abs/1502.02791, 2015.

\bibitem{gtav}
Mark Martinez, Chawin Sitawarin, Kevin Finch, Lennart Meincke, Alex Yablonski,
  and Alain Kornhauser.
\newblock Beyond grand theft auto v for training, testing and enhancing deep
  learning in self driving cars, 2017.

\bibitem{geodesic1}
Pietro Morerio, Jacopo Cavazza, and Vittorio Murino.
\newblock Minimal-entropy correlation alignment for unsupervised deep domain
  adaptation.
\newblock {\em CoRR}, abs/1711.10288, 2017.

\bibitem{mullerpaper}
Matthias M{\"{u}}ller, Alexey Dosovitskiy, Bernard Ghanem, and Vladlen Koltun.
\newblock Driving policy transfer via modularity and abstraction.
\newblock {\em CoRR}, abs/1804.09364, 2018.

\bibitem{mitpaper}
Bowen Pan, Jiankai Sun, Alex Andonian, Aude Oliva, and Bolei Zhou.
\newblock Cross-view semantic segmentation for sensing surroundings.
\newblock {\em CoRR}, abs/1906.03560, 2019.

\bibitem{robustobject}
Cuong~Cao Pham and Jae~Wook Jeon.
\newblock Robust object proposals re-ranking for object detection in autonomous
  driving using convolutional neural networks.
\newblock {\em Signal Processing: Image Communication}, 53:110 -- 122, 2017.

\bibitem{qian2020endtoend}
Rui Qian, Divyansh Garg, Yan Wang, Yurong You, Serge Belongie, Bharath
  Hariharan, Mark Campbell, Kilian~Q. Weinberger, and Wei-Lun Chao.
\newblock End-to-end pseudo-lidar for image-based 3d object detection, 2020.

\bibitem{crossdomainimagegen}
Yaniv Taigman, Adam Polyak, and Lior Wolf.
\newblock Unsupervised cross-domain image generation.
\newblock {\em CoRR}, abs/1611.02200, 2016.

\bibitem{vianney2019refinedmpl}
Jean Marie~Uwabeza Vianney, Shubhra Aich, and Bingbing Liu.
\newblock Refinedmpl: Refined monocular pseudolidar for 3d object detection in
  autonomous driving, 2019.

\bibitem{wang2019monocular}
Dequan Wang, Coline Devin, Qi-Zhi Cai, Philipp Kr{\"a}henb{\"u}hl, and Trevor
  Darrell.
\newblock Monocular plan view networks for autonomous driving.
\newblock {\em arXiv preprint arXiv:1905.06937}, 2019.

\bibitem{hrnet}
Jingdong Wang, Ke Sun, Tianheng Cheng, Borui Jiang, Chaorui Deng, Yang Zhao,
  Dong Liu, Yadong Mu, Mingkui Tan, Xinggang Wang, Wenyu Liu, and Bin Xiao.
\newblock Deep high-resolution representation learning for visual recognition.
\newblock {\em CoRR}, abs/1908.07919, 2019.

\bibitem{pseudolidar}
Yan Wang, Wei-Lun Chao, Divyansh Garg, Bharath Hariharan, Mark Campbell, and
  Kilian Weinberger.
\newblock Pseudo-lidar from visual depth estimation: Bridging the gap in 3d
  object detection for autonomous driving.
\newblock In {\em CVPR}, 2019.

\bibitem{geodesic2}
Yifei Wang, Wen Li, Dengxin Dai, and Luc~Van Gool.
\newblock Deep domain adaptation by geodesic distance minimization.
\newblock {\em CoRR}, abs/1707.09842, 2017.

\bibitem{dcan}
Zuxuan Wu, Xintong Han, Yen{-}Liang Lin, Mustafa~G{\"{o}}khan Uzunbas, Tom
  Goldstein, Ser{-}Nam Lim, and Larry~S. Davis.
\newblock {DCAN:} dual channel-wise alignment networks for unsupervised scene
  adaptation.
\newblock {\em CoRR}, abs/1804.05827, 2018.

\bibitem{pano2cad}
Jiu Xu, Bj{\"{o}}rn Stenger, Tommi Kerola, and Tony Tung.
\newblock Pano2cad: Room layout from {A} single panorama image.
\newblock {\em CoRR}, abs/1609.09270, 2016.

\bibitem{yoo20203d}
Jin~Hyeok Yoo, Yeocheol Kim, Ji~Song Kim, and Jun~Won Choi.
\newblock 3d-cvf: Generating joint camera and lidar features using cross-view
  spatial feature fusion for 3d object detection.
\newblock {\em arXiv preprint arXiv:2004.12636}, 2020.

\bibitem{you2019pseudolidarpp}
Yurong You, Yan Wang, Wei-Lun Chao, Divyansh Garg, Geoff Pleiss, Bharath
  Hariharan, Mark Campbell, and Kilian~Q. Weinberger.
\newblock Pseudo-lidar++: Accurate depth for 3d object detection in autonomous
  driving, 2019.

\bibitem{ganbev}
Xinge Zhu, Zhichao Yin, Jianping Shi, Hongsheng Li, and Dahua Lin.
\newblock Generative adversarial frontal view to bird view synthesis.
\newblock {\em CoRR}, abs/1808.00327, 2018.

\bibitem{layoutnet}
Chuhang Zou, Alex Colburn, Qi Shan, and Derek Hoiem.
\newblock Layoutnet: Reconstructing the 3d room layout from a single {RGB}
  image.
\newblock {\em CoRR}, abs/1803.08999, 2018.

\end{thebibliography}
}

\end{document}